%% file: main.tex
\documentclass[sigconf,authorversion,nonacm]{acmart}

\usepackage{enumitem}

\usepackage{booktabs} 

\usepackage{csquotes}
\usepackage{amsmath}
\usepackage{bm}
\usepackage{outlines}

\usepackage{caption}
\usepackage{subcaption}

\usepackage{siunitx}
\usepackage{graphicx}

\usepackage{pifont}
\usepackage{array}
\newcolumntype{P}[1]{>{\centering\arraybackslash}m{#1}}
\newcolumntype{L}[1]{>{\arraybackslash}m{#1}}

\usepackage{xspace}

\newcommand{\attdmm}{\emph{Att}DMM\xspace}
\newcommand{\KL}{\mathrm{KL}}
\newcommand{\K}{\mathrm{K}}
\newcommand{\ELBO}{\mathrm{ELBO}}
\newcommand{\ReLU}{\mathrm{ReLU}}
\newcommand{\sigmoid}{\mathrm{sigmoid}}
\newcommand{\Softplus}{\mathrm{softplus}}
\newcommand{\const}{\mathrm{const}}

\newcommand\ie{i.\,e.\xspace}
\newcommand\eg{e.\,g.\xspace}

\newcommand\naive{na\"\i ve}

\usepackage[%
capitalize,
sort&compress
]{cleveref}

\usepackage{url}
\usepackage{balance} 

\usepackage{multirow}

\makeatletter
\renewcommand{\fps@figure}{htb}         
\renewcommand{\fps@table}{htb}         
\newcommand*{\rom}[1]{\expandafter\@slowromancap\romannumeral #1@}

\newcommand{\sapsii}{SAPS-\rom{2}\xspace}

\newcommand{\mimiciii}{MIMIC-\rom{3}\xspace}

\makeatother 

\usepackage{bbm}
\usepackage[ruled]{algorithm2e}

\allowdisplaybreaks

\begin{document}
\sloppy

\fancyhead{}

\title[Attentive Deep Markov Model for ICU Risk Scoring]{\attdmm{}: An Attentive Deep Markov Model for Risk Scoring in Intensive Care Units}

\subtitle{}

\author{Yilmazcan \"{O}zyurt}

\affiliation{%
	\institution{ETH Z\"urich}
	\streetaddress{Weinbergstr. 56/58}
	\city{8092 Z\"urich}
	\state{Switzerland}
}
\email{yozyurt@ethz.ch}

\author{Mathias Kraus}

\affiliation{%
	\institution{FAU Erlangen-Nuremberg}
	\streetaddress{Lange Gasse 20}
	\city{90403 Nuremberg}
	\state{Germany}
}
\email{mathias.kraus@fau.de}

\author{Tobias Hatt}

\affiliation{%
	\institution{ETH Z\"urich}
	\streetaddress{Weinbergstr. 56/58}
	\city{8092 Z\"urich}
	\state{Switzerland}
}
\email{thatt@ethz.ch}

\author{Stefan Feuerriegel}
\affiliation{%
	\institution{ETH Z\"urich}
	\streetaddress{Weinbergstr. 56/58}
	\city{8092 Z\"urich}
	\state{Switzerland}
}
\email{sfeuerriegel@ethz.ch}

\begin{abstract}

Clinical practice in intensive care units (ICUs) requires early warnings when a patient’s condition is about to deteriorate so that preventive measures can be undertaken. To this end, prediction algorithms have been developed that estimate the risk of mortality in ICUs. In this work, we propose a novel generative deep probabilistic model for real-time risk scoring in ICUs. Specifically, we develop an attentive deep Markov model called \attdmm{}. To the best of our knowledge, \attdmm is the first ICU prediction model that jointly learns both long-term disease dynamics (via attention) and different disease states in health trajectory (via a latent variable model). Our evaluations were based on an established baseline dataset (\mimiciii{}) with 53,423 ICU stays. The results confirm that compared to state-of-the-art baselines, our \attdmm{} was superior: \attdmm{} achieved an area under the receiver operating characteristic curve (AUROC) of 0.876, which yielded an improvement over the state-of-the-art method by 2.2\,\%. In addition, the risk score from the \attdmm{} provided warnings several hours earlier. Thereby, our model shows a path towards identifying patients at risk so that health practitioners can intervene early and save patient lives.

\end{abstract}



\maketitle

\section{Introduction}

Intensive care units (ICUs) provide healthcare to patients with severe or life-threating illnesses. ICUs receive patients directly from an emergency unit, from other wards if their condition deteriorates rapidly, or after surgery. In ICUs, patients require constant care to ensure normal body functions. Yet, due to the severity of underlying illnesses, their health trajectories cannot always be stabilized. Owing to this, mortality rates in ICUs are among the highest across all hospital units \citep{molina2014outcomes} and are estimated to be 8--19\,\% \citep{mukhopadhyay2014risk}.

To ensure normal body functions, patients in ICUs are subject to extensive monitoring \citep{johnson2016mimic}. Examples are the monitoring of body temperature, heart rate, and blood pressure. Based on this, clinical professionals determine a \textbf{}{risk score} that identifies the probability of in-hospital mortality. The risk score is crucial for decision-making in ICU practice as it guides the treatment plans \citep{lighthall2015understanding, bouch2008severity, nates2016icu}. In addition, it provides early warnings when a health condition is about to deteriorate so that preventive measures can be taken.

The clinical literature indicates the development of several methods for risk scoring in ICUs that are nowadays widely used in clinical practice. Among the most widely applied ones is the simplified acute physiology score (SAPS) \citep{moreno2005saps, metnitz2005saps}, which assesses the severity of the health condition as defined by the probability of patient in-hospital mortality. For this, the risk score makes use of measurements that indicate vital signs, such as body temperature, heart rate, and blood pressure. However, the aforementioned risk scores are computed through overly simple decision rules that operate only on a few measurements from selected timestamps. In other words, the complete time-series of measurements is ignored, and because of this, the prediction power concerning
patient mortality is limited.

Recent works have addressed mortality prediction in ICUs through the use of machine learning. On the one hand, neural networks have been used so that long-term temporal dependencies are captured. Examples of neural networks that have been adapted for ICU predictions are long short-term memory (LSTM) \citep{zheng2018using, ge2018interpretable, thorsen2020dynamic} and gated recurrent unit (GRU) \citep{de2018deep, che2018recurrent}. These models have been powerful in representing complex interactions among (high-dimensional) measurements and thus represent the state of the art. On the other hand, latent variable models have been used for ICU prediction \citep{ghassemi2014unfolding, ghassemi2015multivariate}. In this case, the latent variables allow latent disease states in the health trajectory to be captured. However, we are not aware of any previous works that have combined the strengths of neural networks and latent variables into a joint model for predicting ICU mortality.

\textbf{Proposed model:\footnote{The code is available from https://github.com/oezyurty/AttDMM}} In this work, we propose a novel generative deep probabilistic model for predicting mortality risk in ICUs. Specifically, we develop an attentive deep Markov model called \attdmm{}. Based on this, our model allows us to jointly capture (1)~long-term disease dynamics (via attention) and (2)~different disease states in the health trajectory (via a latent variable model). To the best of our knowledge, our model is the first combination of a deep Markov model with an attention mechanism. In addition, \attdmm{} is closely aligned with the needs in clinical practice providing confidence interval of the real-time risk score that further facilitates decision-making. Finally, we show how to estimate \attdmm{} via an end-to-end training task by a tailored evidence lower bound (ELBO). 
 
\textbf{Results:} Our \attdmm was evaluated on an established baseline dataset from clinical practice, \mimiciii \citep{johnson2016mimic}, comprising 53,423 ICU stays. For each ICU stay, we evaluated the performance across two prediction tasks. First, we predicted the mortality risk from measurements spanning the first 48 hours after ICU admission. This prediction task is analogous to that of risk scoring from the clinical literature \citep[\eg,][]{moreno2005saps, metnitz2005saps,zimmerman2006acute,higgins2007assessing} and thus facilitates comparability with prior literature. In this prediction task, state-of-the-art baselines were consistently outperformed. Compared to the baselines, the proposed \attdmm achieved a performance improvement in the area under the receiver operating characteristic curve (AUROC) by 2.2\,\% and in area under the precision-recall curve (AUPRC) by 2.4\,\%. On top of that, our model achieved the same AUROC (AUPRC) performance as the best baseline 12 hours (6 hours) earlier. Second, we assessed a prediction task in which we make long-term forecasts of mortality risk. This prediction task is demanded by clinical practice as many conditions remain stable for a fairly long time window but afterwards deteriorate suddenly. We found that \attdmm outperforms state-of-the-art baselines in terms of AUROC by 2.2\,\% and AUPRC, remarkably, by 5.4\,\%.

\textbf{Contributions:} Our work advances machine learning for ICU predictions in the following ways:
\begin{enumerate}[leftmargin=15pt]
    \item We present a novel generative deep probabilistic model called \attdmm{}. For this, we combine a deep Markov model with an attention mechanism for risk scoring. To the best of our knowledge, \attdmm{} is the first deep Markov model that is tailored to mortality predictions in ICUs.
    \item Our evaluation demonstrated that \emph{Att}DMM achieved state-of-the-art performance. Specifically, our model outperformed existing baselines by a considerable margin. Thereby, our work enables precise warnings in critical care so that patient lives can be saved.
    \item Our \attdmm{} adheres to the needs of clinical practice in critical care by informing about the precision of the prediction. It provides a confidence interval of the real-time mortality risk score.  Based on it, practitioners can carefully assess their decision-making with regard to the confidence intervals. Thereby, we overcome a shortcoming of existing works in machine learning for ICU predictions, which return only the point estimate of ICU mortality.
\end{enumerate}

\section{Related Work}

\subsection{Machine learning in healthcare}

Machine learning in healthcare has different objectives. Given the breadth of research, we summarize key streams in the following (see \citep{allam2020patient} for a detailed literature review). Examples are predicting hospital readmission risk  \citep[\eg,][]{xiao2018readmission, donze2013potentially, hasan2010hospital, van2020estimating}, forecasting the course of symptoms \citep[\eg,][]{desantis2011hidden, shirley2010hidden, bartolomeo2011progression, maag2021model}, and predicting  diagnosis codes of future diseases \citep[\eg,][]{miotto2016deep, cheng2016risk, choi2016medical, ma2017dipole, choi2016retain, zhang2020inprem, luo2020hitanet}. Depending on the objective of the prediction, different sources of clinical data are used. Typical examples comprise electronic health records, sensor measurements (\eg, in ICUs), and patient sociodemographics (often termed \textquote{risk factors} in a clinical context). The underlying prediction models also differ, for instance, in whether they handle static and/or time-series data,  warrant interpretability, and  model prediction intervals (\ie, whether they output confidence intervals to assess the uncertainty around point estimates) \citep[\eg,][]{ma2017dipole, choi2016retain, zhang2020inprem}.  

The use of machine learning in healthcare allows clinical practitioners to obtain predictions about the current (and future) health condition of a patient. Based on this, clinical practitioners can adapt their treatment plans accordingly (\eg, by planning preventive interventions or choosing different treatments).

\subsection{Machine learning in intensive care units } 

\textbf{Risk prediction in ICUs:} Machine learning in ICU settings has different objectives, such as predicting adverse events like sepsis \citep[\eg,][]{kam2017learning, scherpf2019predicting, sheetrit2019temporal, kaji2019attention, yin2020identifying}, while a predominant focus in the literature is on predicting mortality risk. For this, one uses various measurements of vital signs, such as body temperature, heart rate, and blood pressure. They are widely regarded as important indicators that describe a patient's health status in critical care \citep{le1993new, ferreira2001serial}. 

In practice, there is variability in which measurements are recorded for a specific patient depending on her condition and reason of hospitalization. For instance, for some diseases, sodium and potassium levels are highly indicative of the future course, while for COVID-19, a focus might be placed on measuring respiratory-related thoracic movements. As such, some measurements might not be subject to recording (due to the setting) and, hence, are missing for the complete (or a partial) ICU stay. Hence, the missingness of ICU measurements must be appropriately modeled in the context of ICU prediction \citep[\eg,][]{che2018recurrent}. 

\textbf{Clinical practice in ICU risk scoring:} In clinical practice, predictions of mortality risk are computed through simple decision rules. A key benefit of decision rules from clinical practice is that they directly output a dynamic \emph{risk score}. Common examples are the so-called simplified acute physiology score (SAPS) \citep{moreno2005saps, metnitz2005saps}, the acute physiology and chronic health evaluation (APACHE)  \citep{zimmerman2006acute}, and mortality probability model (MPM) \citep{higgins2007assessing}. To allow for straightforward use, the decision rules rely only upon a few sensor measurements and, in particular, additional characteristics of patients (\ie, risk factors) are not considered. In essence, these decision rules determine the mortality risk through a linear combination of different vital signs. However, only the most critical value from the last 48 hours is considered, rather than the complete time-series of vital signs. Hence, the rich information embedded in high-resolution measurements is---to a large extent---ignored which limits the prediction power. 

\textbf{Machine learning for ICU mortality risk scoring:} Recent works have approached ICU mortality prediction using machine learning. For this, several sequential neural networks have been adapted specifically to ICU settings, namely long short-term memory networks \citep{ge2018interpretable, harutyunyan2017multitask} and gated recurrent units \citep{che2018recurrent, de2018deep, shukla2018modeling}. For ICU predictions, state-of-the-art sequential networks are designed to capture long-term dependencies in high-resolution time-series. However, existing machine learning does not explicitly model different disease states in trajectory, which would require a latent variable approach.

\subsection{Latent variable models} 

Latent variable models have the ability to account for the fact that patient health trajectories undergo different disease states (\eg, acute, stable). However, latent variable models explicitly consider the fact that such disease states cannot be directly observed and must thus be treated as latent \citep[\eg,][]{desantis2011hidden, shirley2010hidden, bartolomeo2011progression}. In principle, such models assume that the relation between health measurements and the disease states are stochastic. A na{\"i}ve example is a hidden Markov model. However, these models are generally made for modeling a sequence of observations instead of classification tasks. 

To adapt latent variable models for prediction tasks, some works \citep{alaa2017individualized, yoon2016forecasticu, alaa2017personalized} encode the risk into the latent variable and adopt sequential hypothesis testing \citep{wald1945sequential}. Others \citep{ghassemi2014unfolding, ghassemi2015multivariate} feed latent variables into another prediction algorithm, while even others \citep{meng2019hierarchical, yu2014real, martino2020multivariate} match discrete latent variables to the risk profiles. However, these approaches have only a limited ability in capturing long-term temporal dynamics, rely on discrete latent variables, and assume linearity in the transition and emission components. 

The deep Markov model (DMM) \citep{krishnan2017structured} overcomes the above limitations by introducing continuous latent variables, non-linear transition networks, and non-linear emission networks. However, the DMM presents an unsupervised framework, because of which an application of the DMM to the prediction of mortality risk is precluded. Instead, a new model is needed, and for this reason, we develop an attentive deep Markov model to address the above mentioned prediction tasks.

\textbf{Research gap:} To the best of our knowledge, no prior work has combined the strength of neural networks and latent variable approaches for the purpose of ICU prediction. To fill this gap, we propose a novel attentive deep Markov model called \attdmm{}.

\section{\attdmm{}: Attentive Deep Markov Model for ICU Risk Scoring } \label{sec:attdmm}

\subsection{Problem statement}

In the following, we develop \attdmm{}, which processes ICU stays in order to predict the ICU mortality label $y \in \{0,1\}$. A label $y=0$ denotes a discharge from ICU, whereas $y=1$ denotes that the patient has died during the ICU stay.

The input to our model are ICU stays. Each stay is represented by static and time-series features: (1)~Static features are denoted by $s$. These encode potential risk factors at the patient level (\eg, gender, age) and thus describe the between-patient heterogeneity. (2)~Time-series features are denoted by $\{x_t\}_{t=1}^T$. These encode a high-resolution time-series with various measurements for each time step $t = 1, \ldots, T$ throughout the ICU stay (\eg, in \mimiciii, they are sampled at a 2-hour resolution). Some measurements might not be subject to recording, and, therefore, are missing for the complete (or a partial) ICU stay. To address this issue, we first decompose $x_t$ into $\{x_{ti}\}_{i=1}^M$, where $M$ is the number of features (\ie, types of measurements). Then, for each $x_{ti}$, we derive the mask variable $m_{ti}$ that denotes if $x_{ti}$ will be imputed ($m_{ti} = 0$) or not ($m_{ti} = 1$).

\subsection{Specification of \attdmm} \label{sec:sub_attdmm}

\attdmm is a generative deep probabilistic model for predicting mortality risk in ICUs. The model takes lab measurements recorded during an ICU stay as input. The measurements are processed to capture the latent disease state in a patient trajectory. This is achieved via a latent variable model which captures the latent disease state at time $t$ by a latent variable $z_t$. Based on the series of latent variables, our model infers the ICU mortality risk of a patient. This is achieved via an attention mechanism. 

\begin{figure}[h]
    \centering
    \includegraphics[width=0.6\columnwidth]{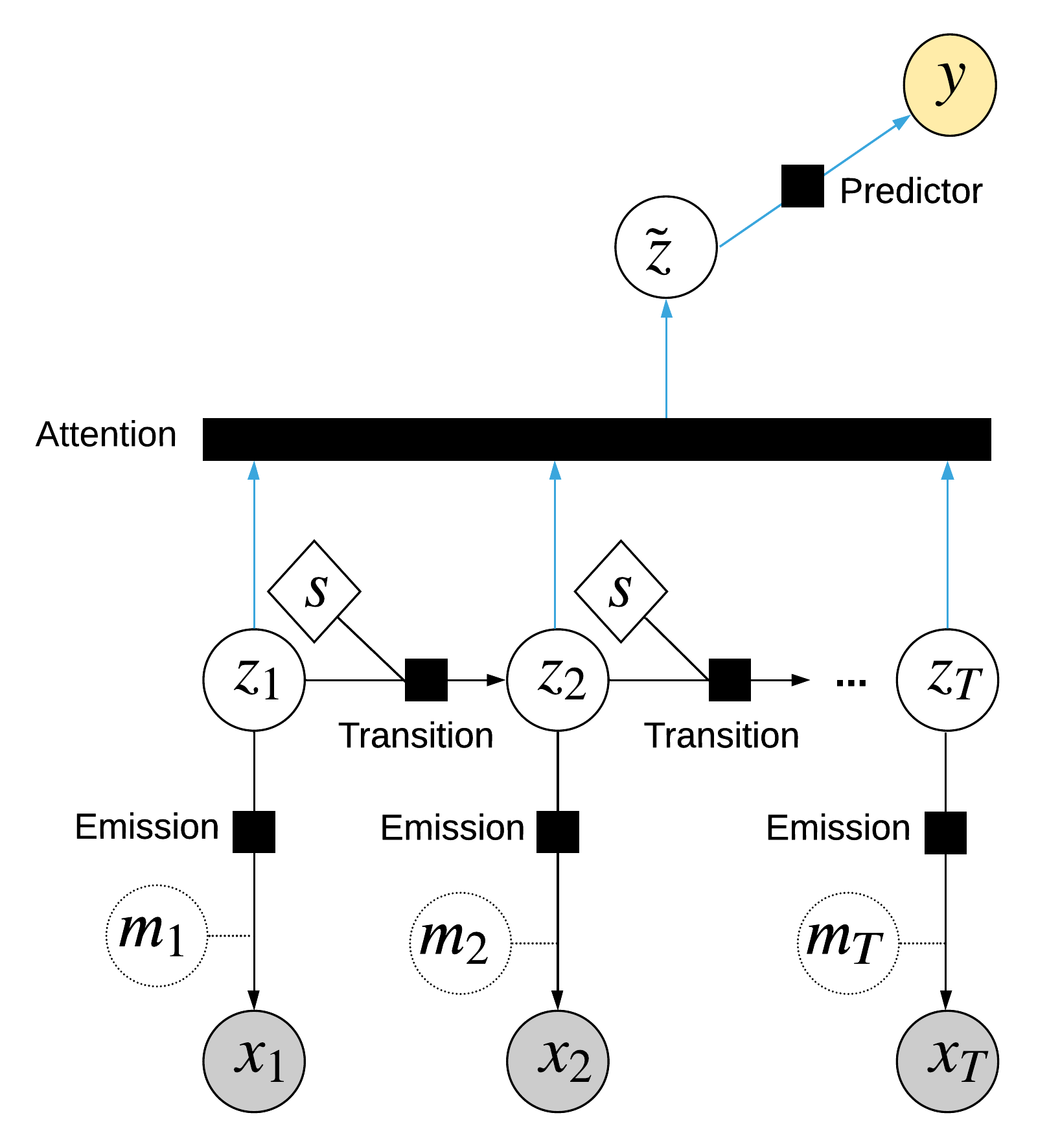}
    \caption{\attdmm. Black squares denote neural networks.}
    \label{fig:attdmm_generative_model}
\end{figure}

\attdmm has 4 model components, which are depicted in Figure~\ref{fig:attdmm_generative_model}. A \textbf{(1)~transition network} specifies the transition probability among consecutive latent variables (\ie, $z_{t-1}$ and $z_t$). It further accommodates between-patient heterogeneity. An \textbf{(2)~emission network} models the probability of a lab measurement given the latent variable from a certain time step. In this case, it adheres to the assumption that the lab measurements are stochastically linked to the latent variables. The combination of a transition network and emission network represents the generative part of the model, which is used as an ELBO regularization in the loss function. An \textbf{(3)~attention network} outputs the summary representation of the patient trajectory $\Tilde{z}$, based on all latent variables $z_{1:T}$. Finally, a \textbf{(4)~predictor network} takes this summary representation as input and predicts the ICU mortality risk of a patient. 

In terms of notation, let $\ReLU$ denote the rectified linear unit, $\odot$ denote the element-wise multiplication, and $[.;\,.]$ denote the concatenation of two vectors.

\subsection{Model components}

\textbf{(1)~Transition network:} This component specifies the transition probability among consecutive latent variables. For this, we build upon the Markov property; \ie, the current latent variable only depends on the previous one. Further, it accommodates the between-patient heterogeneity by making use of the patient's static variable $s$. 

Formally, the transition network models the distribution of a latent variable $z_t$ conditional on the inputs $z_{t-1}$ and $s$, denoted by $p(z_t \,\mid\, z_{t-1}, \, s)$. This distribution is parametrized by a multivariate Gaussian distribution with a diagonal covariance, denoted by $\mathcal{N}(\mu_{z_t}, \, \Sigma_{z_t})$, where $\Sigma_{z_t}$ is the covariance matrix with $\sigma_{z_t}^2$ on the diagonal and 0 otherwise. The mean $\mu_{z_t}$ is a sum of two terms: (1) a linear contribution of the inputs, denoted by $\bar{\mu}_{z_t}$, and (2) a non-linear contribution of the inputs, denoted by $\Tilde{\mu}_{z_t}$. The weights of these two terms are controlled by a gate $g_{z_t}$. The diagonal $\sigma_{z_t}^2$ is derived from $\Tilde{\mu}_{z_t}$ (plus a constant\footnote{A constant term is added to the diagonal covariance to ensure stability in the ELBO computation. This applies to all Gaussian distributions introduced in \attdmm.}). The mathematical formulation of the transition network is as follows:
\begin{gather}
    g_{z_t}^{\prime} = \ReLU(W^{t}_{g^\prime}\, [z_{t-1};\,s] + b^{t}_{g^{\prime}}), \\
    g_{z_t} = \sigmoid (W^{t}_{g}\, g_{z_t}^{\prime} +  b^{t}_g), \\
    \Tilde{\mu}_{z_t}^{\prime} = \ReLU(W^{t}_{\Tilde{\mu}^{\prime}}\, [z_{t-1};\,s] + b^{t}_{\Tilde{\mu}^{\prime}}), \\
    \Tilde{\mu}_{z_t} = W^{t}_{\Tilde{\mu}}\, \Tilde{\mu}_{z_t}^{\prime} + b^{t}_{\Tilde{\mu}}, \\
    \bar{\mu}_{z_t} = W^{t}_{\bar{\mu}}\, [z_{t-1};\,s] + b^{t}_{\bar{\mu}}, \\
    \mu_{z_t} = g_{z_t} \odot \Tilde{\mu}_{z_t} + (1-g_{z_t}) \odot \bar{\mu}_{z_t}, \\
    \sigma_{z_t} = \Softplus (W^{t}_{\sigma}\, \ReLU( \Tilde{\mu}_{z_t}) + b^{t}_{\sigma}) + \const. , \\
   z_t \sim p(z_t \,\mid\, z_{t-1}, \, s) = \mathcal{N}(\mu_{z_t}, \, \Sigma_{z_t})
\end{gather}
with matrices $W^{t}_{g^\prime}$, $W^{t}_{g} $, $W^{t}_{\Tilde{\mu}^{\prime}}$, $W^{t}_{\Tilde{\mu}}$, $W^{t}_{\bar{\mu}}$, and $W^{t}_{\sigma}$ and bias vectors $b^{t}_{g^{\prime}}$, $b^{t}_g$, $b^{t}_{\Tilde{\mu}^{\prime}}$, $ b^{t}_{\Tilde{\mu}}$, $b^{t}_{\bar{\mu}}$, and $b^{t}_{\sigma}$. 

\textbf{(2)~Emission network:} The emission network outputs the probability of a lab measurement given the latent variable. Formally, the emission network models the distribution of $x_t$ given the latent variable $z_t$, denoted by $p(x_t \,\mid\, z_t)$. In the distribution, each feature $x_{ti}$ of $x_t$ is modeled as conditionally independent from each other, \ie,
\begin{equation}
    p(x_t \,\mid\, z_t) = \prod_{i : m_{ti}=1} p(x_{ti} \,\mid\, z_t),
\end{equation}
where $m_{ti}$ is leveraged to mask out unobserved entries. The distribution $p(x_{ti} \,\mid\, z_t)$ is parametrized by a univariate Gaussian distribution, denoted by $\mathcal{N}(\mu_{x_{ti}}, \, \sigma_{x_{ti}}^2)$, where $\mu_{x_{ti}}$ and $\sigma_{x_{ti}}^2$ are the $i$-th indices of the vectors $\mu_{x_t}$ and $\sigma_{x_t}^2$.  With the same model assumptions from the transition network, the mean $\mu_{x_t}$ is a sum of two terms: (1) a linear contribution of the inputs, denoted by $\bar{\mu}_{x_t}$, and (2) a non-linear contribution of the inputs, denoted by $\Tilde{\mu}_{x_t}$. The weights of these two terms are controlled by a gate $g_{x_t}$. The variance $\sigma_{x_t}^2$ is derived from $\Tilde{\mu}_{x_t}$ (plus a constant as above). The mathematical formulation of the emission network is as follows:
\begin{gather}
    g_{x_t}^{\prime} = \ReLU(W^{e}_{g^\prime}\, z_{t} + b^{e}_{g^\prime}) , \\
    g_{x_t} = \sigmoid (W^{e}_{g}\, g_{x_t}^{\prime} +  b^{e}_{g}) , \\
    \Tilde{\mu}_{x_t}^{\prime} = \ReLU(W^{e}_{\Tilde{\mu}^{\prime}}\, z_{t} + b^{e}_{\Tilde{\mu}^{\prime}}) , \\
    \Tilde{\mu}_{x_t} = W^{e}_{\Tilde{\mu}}\, \Tilde{\mu}_{x_t}^{\prime} + b^{e}_{\Tilde{\mu}} , \\
    \bar{\mu}_{x_t} = W^{e}_{\bar{\mu}}\, z_{t} + b^{e}_{\bar{\mu}} , \\
    \mu_{x_t} = g_{x_t} \odot \Tilde{\mu}_{x_t} + (1-g_{x_t}) \odot \bar{\mu}_{x_t} , \\
    \sigma_{x_t} = \Softplus (W^{e}_{\sigma}\, \ReLU( \Tilde{\mu}_{x_t}) + b^{e}_{\sigma}) + \const. , \\
   x_{ti} \sim p(x_{ti} \,\mid\, z_{t}) = \mathcal{N}(\mu_{x_{ti}}, \, \sigma_{x_{ti}}^2), \ \ \forall i \in \{1,\,...,\,M\}
\end{gather}
with matrices $W^{e}_{g^\prime}$, $W^{e}_{g} $, $W^{e}_{\Tilde{\mu}^{\prime}}$, $W^{e}_{\Tilde{\mu}}$, $W^{e}_{\bar{\mu}}$, and $W^{e}_{\sigma}$ and bias vectors $b^{e}_{g^{\prime}}$, $b^{e}_g$, $b^{e}_{\Tilde{\mu}^{\prime}}$, $ b^{e}_{\Tilde{\mu}}$, $b^{e}_{\bar{\mu}}$, and $b^{e}_{\sigma}$. 

\textbf{(3)~Attention network:} The attention network allows the model to assign a different importance to each latent variable when making inferences of mortality risk. The importance is computed via similarity between a query vector and the latent variable. Thereby, the attention network aggregates the latent variables based on their importance. The resulting vector is used as a summary representation of the patient health trajectory, which is fed into a predictor network. Although this overcomes the Markov property, it allows the model to capture long-term dependencies.

Formally, the attention network outputs the aggregation of latent variables $\tilde{z}$ based on the input sequence  $z_{1:T}$. This is formalized via 
\begin{gather}
    z_t^{\prime} = W_{z^{\prime}}\, z_t + b_{z^{\prime}} ,  \\
    \gamma_t = \frac{\exp(\zeta\, \K[v,\, z_t^{\prime}])}{\sum_{t'=1}^T \exp(\zeta\, \K[v,\, z_{t'}^{\prime}])} , \\
    \tilde{z} = \sum_{t=1}^T \gamma_t\, z_t 
\end{gather}
with matrix $W_{z^{\prime}}$, bias vector $ b_{z^{\prime}}$, query vector $v$, and scalar $\zeta$, where $\K(\cdot,\cdot)$ defines the cosine similarity, \ie, 
\begin{equation}
    \K(v,\,z_t^{\prime}) = \frac{v \cdot z_t^{\prime}}{||v|| \cdot ||z_t^{\prime}||}
\end{equation}

\textbf{(4)~Predictor network:} The predictor network is the final step of making the inference of mortality risk. For this, it predicts $y$ based on $\tilde{z}$ from the attention network. This yields
\begin{equation}
   \hat{y} = \sigmoid(U_y\, \ReLU(W_y\, \tilde{z} + b_y) + c_y) 
\end{equation}
with matrices $W_y$ and $U_y$ and bias vectors $b_y$ and $c_y$. $\hat{y}$ is the output of \attdmm. 

\subsection{Posterior approximation}

\begin{figure}[h]
    \centering
    \includegraphics[width=0.4\columnwidth]{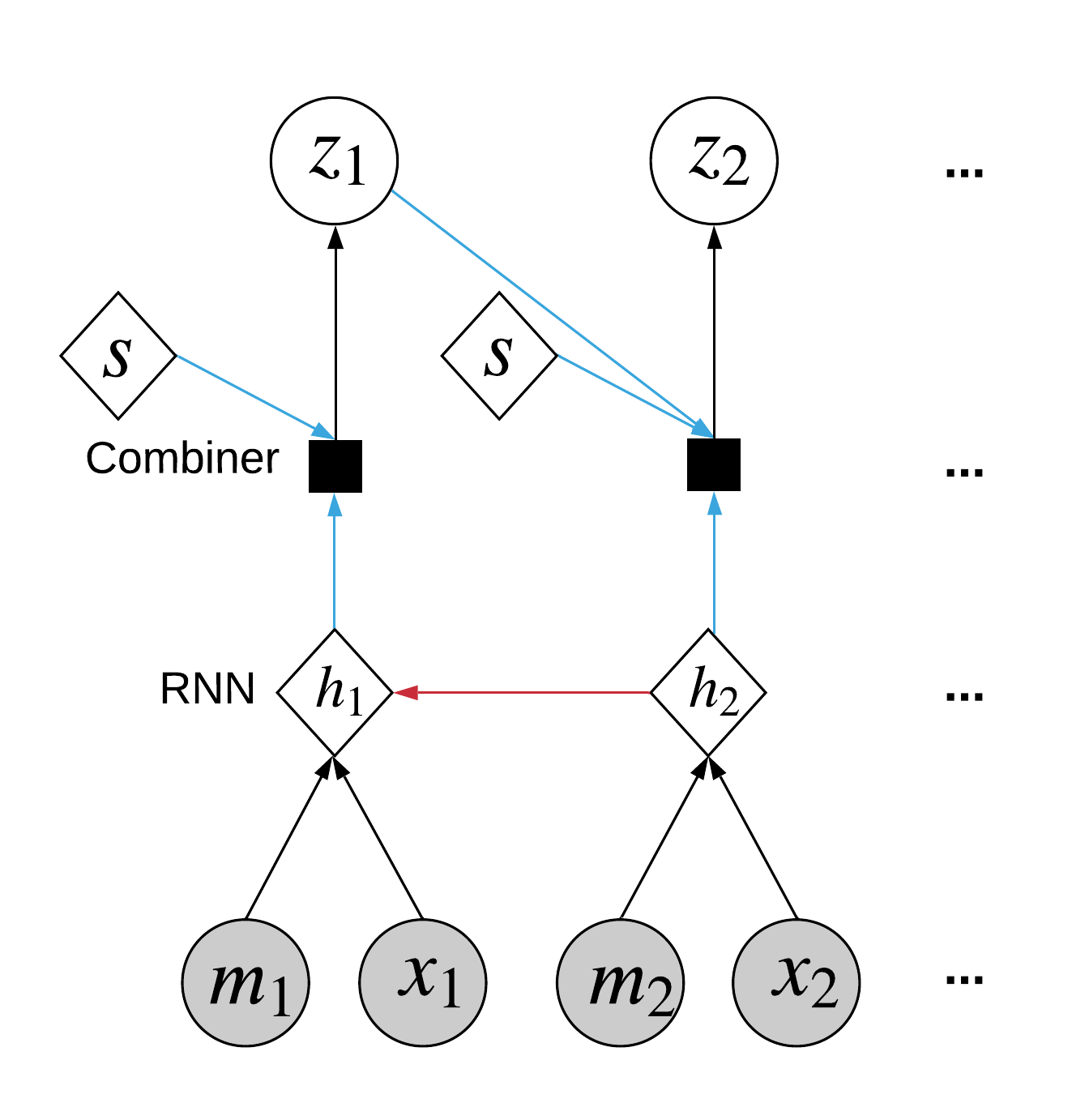}
    \caption{Posterior approximation of \attdmm, rolled-out across time steps. Black squares denote neural networks.}
    \label{fig:attdmm_posterior_approx}
\end{figure}

The computational complexity present in \attdmm hinders the exact inference of the latent variables. Because of this, we develop a posterior approximation that builds upon stochastic variational inference to estimate latent variables (see Figure~\ref{fig:attdmm_posterior_approx}).

For \attdmm, we approximate the posterior distribution of latent variables (denoted as $q$) based on the sequence of lab measurements and  static variables. It further makes use of the missingness of each lab measurement (\ie, masks).

Formally, the approximated posterior distribution $q$ of $z_t$ is based on the inputs $z_{t-1}$, $s$, $x_{t:T}$, and $m_{t:T}$. For this, we proceed as follows. First, we use a recurrent neural network (RNN) for encoding the information carried by the future lab measurements and their missingness, \ie, a concatenation of $x_{t:T}$ and $m_{t:T}$. This yields the encoded information $h_t$. This is formalized via
\begin{gather}
    h_t = \ReLU(W_h\, [m_t;\,x_t] + U_h\, h_{t+1} + b_h) ,
\end{gather}
with matrices $W_h$, $U_h$ and bias vector $b_h$. Second, a combiner network is applied to the inputs of $z_{t-1}$, $s$, and $h_t$. In this case, $z_{t-1}$ and $h_t$ encapsulate the past and future lab measurements. In addition, $s$ is leveraged to accommodate the between-patient heterogeneity in the posterior approximation.  

The combiner network models the approximated distribution of $z_t$, denoted by $q(z_t \,\mid\, z_{t-1},\, s,\, x_{t:T},\, m_{t:T})$. This distribution is parametrized by a multivariate Gaussian distribution with a diagonal covariance, denoted by $\mathcal{N}(\mu_{z_t}, \, \Sigma_{z_t})$, where $\Sigma_{z_t}$ is the covariance matrix with $\sigma_{z_t}^2$ on the diagonal and 0 otherwise. The mathematical formulation of the combiner network is given by
\begin{gather}
    c_t = W_c\, [z_{t-1};\, s] + b_c , \\
    \Tilde{h}_t = 0.5\, \tanh{(c_t + h_t)} , \\
    \mu_{z_t} = W_\mu\, \Tilde{h}_t + b_\mu , \\
    \sigma_{z_t} = \Softplus(W_\sigma\, \Tilde{h}_t + b_\sigma ) + \const. , \\
    z_t \sim q(z_t \,\mid\, z_{t-1},\, s,\, x_{t:T},\, m_{t:T}) = \mathcal{N}(\mu_{z_t},\, \Sigma_{z_t})
\end{gather}
with matrices $W_c$, $W_\mu$, and $W_\sigma$ and bias vectors $b_c$, $b_\mu$, and $b_\sigma$. Altogether, this approximated the posterior of the latent variables of \attdmm, which is leveraged to optimize the ELBO via the stochastic variational inference. See Appendix~\ref{sec:estimation_procedure} for the details of the estimation procedure.

\subsection{Inferring risk scores}

\attdmm{} estimates the probability of in-hospital mortality at time $T$ as follows. The static features $s$, the lab measurements $x_{1:T}$, and the masking of measurements $m_{1:T}$ are fed into the posterior approximation network. Based on it, \attdmm constructs the posterior distribution $q(z_t \,\mid\, z_{t-1},\, s,\, x_{t:T},\, m_{t:T})$, from which the latent variables $z_{1:T}$ are sampled sequentially. This sampling procedure is repeated $N$ times, where the number $N$ is defined by the user. At the end of the sampling procedure, the set of sampled latent variables $\{z_{1:T}^n\}_{n=1}^N$ is processed sequentially by the attention network and the predictor network. \attdmm{} then produces $N$ samples of in-hospital mortality prediction $\{\hat{y}^n\}_{n=1}^N$, whose mean is used for the estimate of the probability of in-hospital mortality.

The complete hospital stay of an ICU patient is tracked by our \attdmm{}, thereby producing the probability of in-hospital mortality over time. In this way, \attdmm{} can inform the physicians in the ICU about the risk scores of each patient. 

\attdmm{} yields a full posterior distribution of the prediction (rather than a point estimate). This distribution is acquired by the multiple samples of the mortality prediction $\{\hat{y}^n\}_{n=1}^N$. This allows us to compute confidence interval of the prediction in order to quantify the uncertainty of our prediction. and, in practice, one can use it in order to decide whether the prediction is sufficiently reliable.

\section{Experimental Setup}

\subsection{Dataset}

Our evaluation was based on an established reference dataset with ICU measurements, namely, \mimiciii{} \citep{johnson2016mimic}. \mimiciii{} is one of the largest publicly available ICU datasets, comprising 38,597 distinct patients and a total of 53,423 ICU stays. Because of this, \mimiciii{} has been used extensively in prior literature for benchmarking \citep[\eg,][]{kam2017learning, scherpf2019predicting, sheetrit2019temporal, kaji2019attention, purushotham2017benchmark, harutyunyan2017multitask, che2018recurrent, ge2018interpretable, de2018deep}. For each ICU stay, the dataset includes a time-series of different measurements, such as temperature, heart rate, and blood pressure. In addition, the dataset also reports sociodemographic variables that describe the heterogeneity among patients, such as age and admission type (\eg, scheduled surgery, unscheduled surgery). The available features for prediction are reported in Appendix~\ref{sec:app_mimic}. For details on the dataset, we refer the reader to \citep{johnson2016mimic}.

We follow the preprocessing pipeline established by prior literature \citep{purushotham2017benchmark, che2018recurrent}. We first remove ICU stays which are shorter than 2 days or longer than 30 days, and ICU stays of patients younger than 15. Further, we only consider the first ICU stay of patients with multiple stays. Afterwards, we applied further preprocessing steps consistent with the aforementioned references \citep{purushotham2017benchmark, che2018recurrent}. For each ICU stay, we extracted features that were sampled regularly (\ie, every 2 hours), and we filled missing values with a forward-backward linear imputation.

The preprocessed dataset contains 31,895 ICU stays. Out of them, the share of ICU stays with in-hospital mortality amounted to 3,311 stays (\ie, 10.38\,\%). The remaining 28,584 stays (\ie 89.62\,\%) resulted in a discharge from the ICU. The length of hospital stay after ICU admission had a mean of 189.21 hours (with a standard deviation of 136.44 hours). Of all the ICU stays, 58.78\,\% required a hospital stay of longer than five days. The distribution of hospitalization length is shown in the Appendix.

\subsection{Prediction tasks}

\subsubsection{Task 1: Mortality prediction (48 hours after ICU admission)}

In this task, we predict the in-hospital mortality based on the measurement data from the first 48 hours after ICU admission. This represents the reference task in the literature for calibrating and evaluating mortality predictions in ICUs. To examine how the prediction performance changes throughout an ICU stay, we further report the prediction performance at different time periods after the ICU admission, ranging from 12 to 48 hours after the ICU admission. For this, all the predictions have only access to the respective (limited) time frame of ICU measurements. This allows us to assess how precise predictions are at the beginning of the ICU stay.

\subsubsection{Task 2: Mortality prediction (complete hospital stay after ICU admission)}

A second prediction task is used to evaluate ICU mortality during the complete hospital stay. Typical use cases are diseases in which the health trajectory remains stable for a large amount of time and, only later, deteriorates suddenly. The relevance of the second prediction task is also seen in the summary statistics of our dataset, in which hospital stays with a duration of more than 48 hours are highly common and account for more than 80\,\% of all ICU admissions. 
 
Formally, we now use the complete time-series of the ICU data during both training and evaluation. We then report the prediction performance in intervals of 2 hours. In this case, we label time with respect to hours to discharge/death. We report the prediction performance for all lead times between $-120$ hours and $0$ hours. Longer lead times were discarded, as there was not sufficient data.
 
We later also report an overall performance score. For this, we aggregate the prediction performance across all time-steps via a weighted micro-level average. Thereby, we weigh each time step by the number of available patients for the evaluation at that time step.

\subsection{Performance metrics} 
 
We compare our \attdmm with other baselines based on two performance metrics. (1)~We report area under the receiver operating characteristic curve (\textbf{AUROC}). This was also chosen in prior literature \citep{che2018recurrent, ge2018interpretable, de2018deep}. We tested whether the improvement in AUROC is statistically significant via DeLong’s test \citep{delong1988comparing}. (2)~We further report area under the precision-recall curve (\textbf{AUPRC}). The AUPRC is frequently used in the clinical literature as it focuses on the performance relative to detecting a negative event (\ie, mortality).

\subsection{Baselines}

We compare our \attdmm{} against an extensive series of baselines that have been carefully crafted for mortality prediction in ICU settings. We first implemented \textbf{\sapsii{}} representing current practice \citep{le1993new}.\footnote{We also considered the use of APACHE and MPM; however, these are known to rely upon features that are not available as part of the preprocessed \mimiciii dataset. Hence, we followed prior literature and compared our model against \sapsii as a baseline from clinical practice.} Formally, \sapsii provides predefined decision rules that make use of the most critical lab measurements of the last 48 hours.

In addition, we implemented \naive{} baselines for performance comparisons. These are a multilayer perceptron (\textbf{MLP}) and a random forest (\textbf{RF}). Both are fed with summary statistics (mean, max, min, std. dev.) of the same features used by current practice. Hence, the \naive{} baselines are intended to reflect the power of the prediction heuristics from clinical practice.

We further adapt latent variable models to ICU mortality prediction. A hidden Markov model (\textbf{HMM}) is utilized with two hidden states, referring to discharge and mortality. Additionally, we use an HMM with multiple hidden states and feed these states into a long short-term memory, denoted as \textbf{HMM+LSTM}. Finally, we crafted a \textbf{DMM} for the prediction task, in which the last latent variable is fed into the predictor network (the same network of the \attdmm).  

We use the following ICU-specific state-of-the-art models: a long short-term memory network for ICU prediction (\textbf{ICU/LSTM}) \citep{ge2018interpretable}; a gated recurrent unit for ICU prediction (\textbf{ICU/GRU}) \citep{de2018deep}; and a gated recurrent unit with decay mechanism for ICU prediction (\textbf{ICU/GRU-D}) \citep{che2018recurrent}. These models can, by nature of the underlying model, handle sequential input of variable-length and are, thus, trained with the same data as \attdmm. We use implementations analogous to those in the prior literature that have been tailored to handle the specific time-series structure in ICU settings (\eg, with regard to missing values, sampling frequency). Appendix~\ref{sec:implementation_details} describes details about the implementation and evaluation of all baseline models as well as \attdmm.

\section{Results}

\subsection{Task 1: Mortality prediction (48 hours after ICU admission)} \label{sec:experiments_48hours}

Table \ref{tab:experiments_48hours} lists the results of in-hospital mortality prediction based on the measurement data from the first 48 hours after ICU admission. The best baseline is given by the ICU/GRU-D with an AUROC of 0.857. In contrast, \attdmm{} achieves an AUROC of 0.876, which is an improvement over the best baseline by 2.2\,\%. The improvement is statistically significant (\emph{p}<0.001). Similarly, \attdmm{} achieved the highest AUPRC with 0.465. Compared to the best baseline (ICU/GRU-D with an AUPRC of 0.454), this is an improvement of 2.4\,\%.

\input{tables/experiments/experiments_48hours}

\begin{figure}[h]
    \begin{subfigure}{1.\columnwidth}
        \centering
        \includegraphics[width=1.\columnwidth]{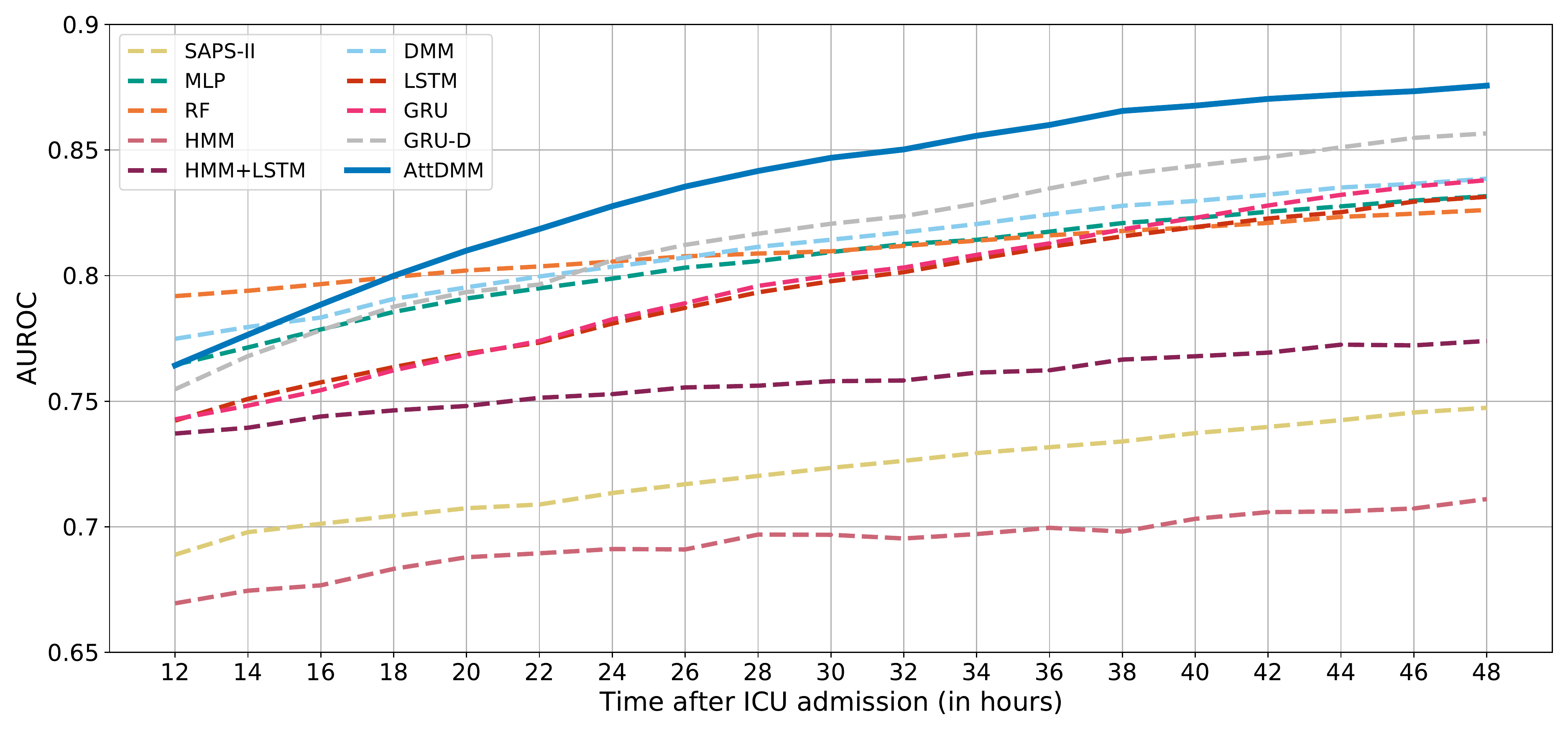}
        \label{fig:task1_auroc}
    \end{subfigure}
    
    \begin{subfigure}{1.\columnwidth}
        \centering
        \includegraphics[width=1.\columnwidth]{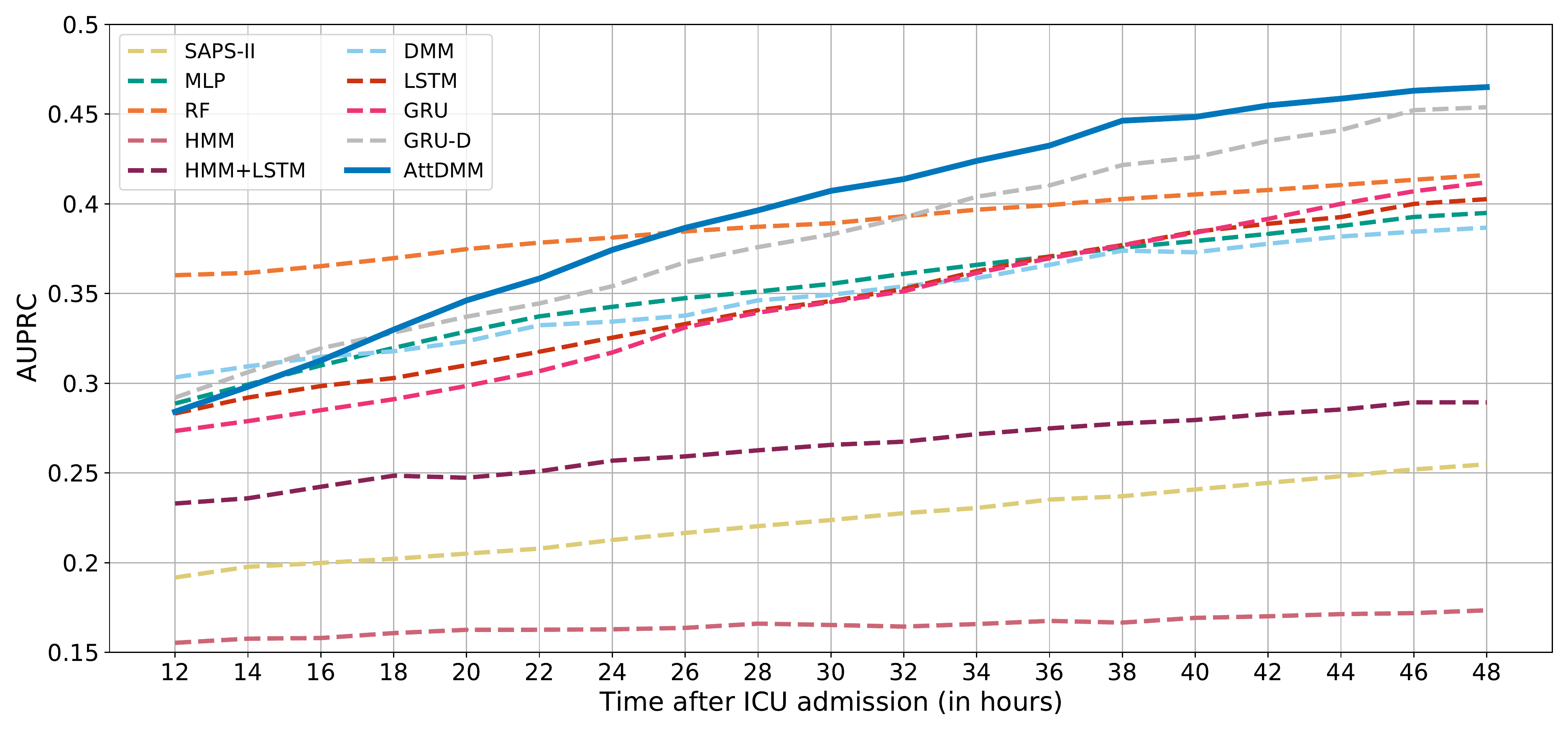}
        \label{fig:task1_auprc}
    \end{subfigure}
    
    \caption{Sensitivity of prediction performance across varying time periods after ICU admission. Shown are two performance metrics: AUROC (top) and AUPRC (bottom).}
    \label{fig:task1}
    
\end{figure}

We further provide a sensitivity analysis in which we study how the prediction performance varies after ICU admission. For this, Figure \ref{fig:task1} depicts the evolution of both the AUROC and AUPRC from 12 to 48 hours (in steps of 2 hours) after ICU admission. All the baselines are outperformed by the proposed \attdmm{}. The improvement of \attdmm{} over the baselines is particularly pronounced for time periods after the first 24 hours. We further compare the models in their minimum time needed after the ICU admission in order to achieve a specific prediction performance. To achieve an AUROC of 0.85, the ICU/GRU-D needs 44 hours of patient data, whereas the same AUROC is achieved by \attdmm{} 12 hours earlier. For AUPRC, the /GRU-D performs better shortly after ICU admission, but after 18 hours, it is outperformed by \attdmm{}. An AUPRC of 0.45 is achieved by the ICU/GRU-D 46 hours after ICU admission, whereas the same AUROC is achieved by \attdmm{} after 40 hours, \ie, 6 hours earlier. 

\subsection{Task 2: Mortality prediction (complete hospital stay after ICU admission)} \label{sec:experiments_cropped_t}

Table \ref{tab:experiments_cropped_t} lists the results of in-hospital mortality prediction based on the complete hospital stay. The best baseline is given by the random forest with an AUROC of 0.846. In contrast, \attdmm{} achieves an AUROC of 0.865. Therefore, \attdmm yields an improvement over the best baseline by 2.2\,\%. The improvement is statistically significant (\emph{p}<0.001). For the AUPRC, our proposed model again achieves the best score amounting to 0.545. In this case, \attdmm{} yields an improvements over the best baseline (ICU/GRU-D with an AUPRC of 0.517) by 5.4\,\%. 
\input{tables/experiments/experiments_cropped_t}

\begin{figure}[h]
    \begin{subfigure}{1.\columnwidth}
        \centering
        \includegraphics[width=1.\columnwidth]{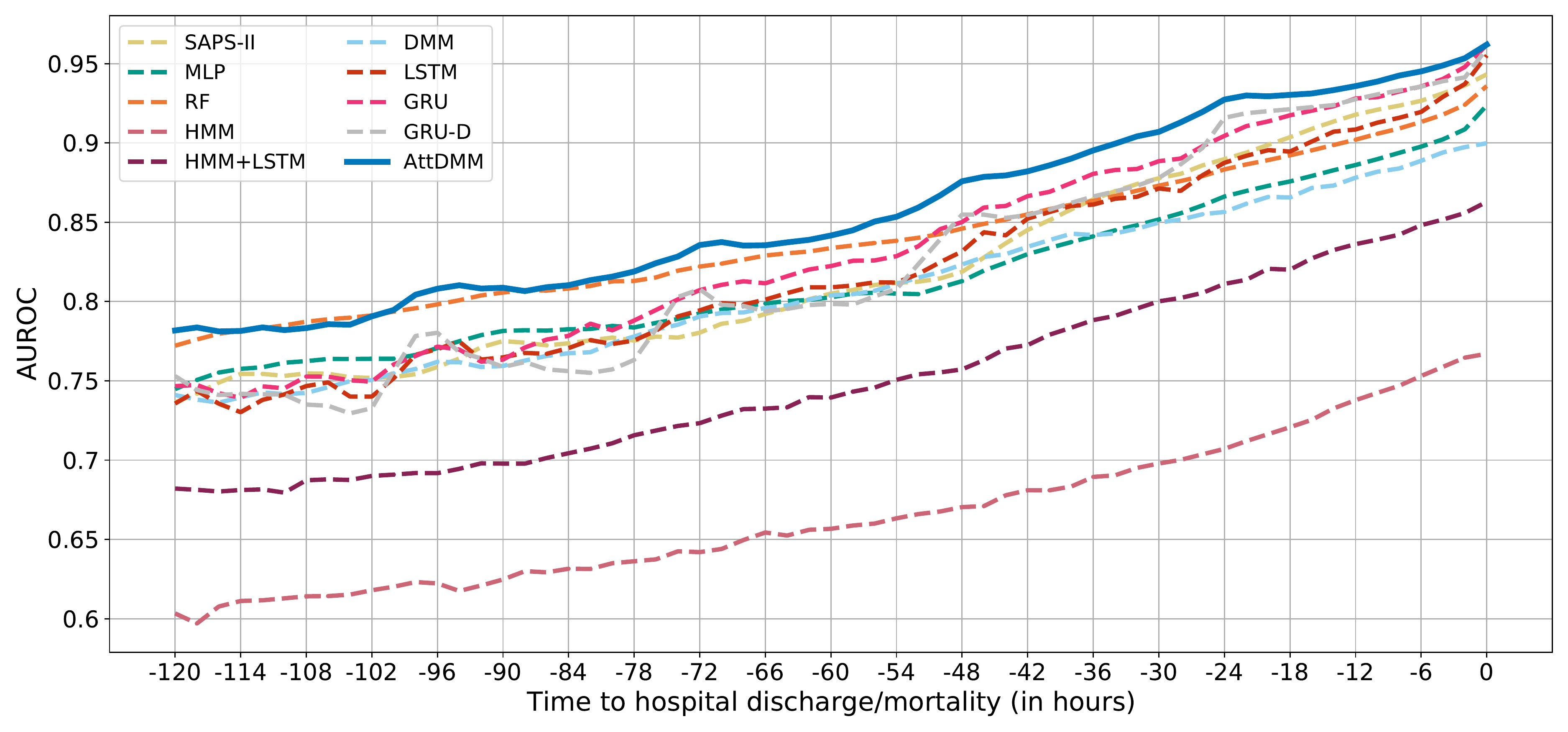}
        \label{fig:task2_auroc}
    \end{subfigure}

    \begin{subfigure}{1.\columnwidth}
        \centering
        \includegraphics[width=1.\columnwidth]{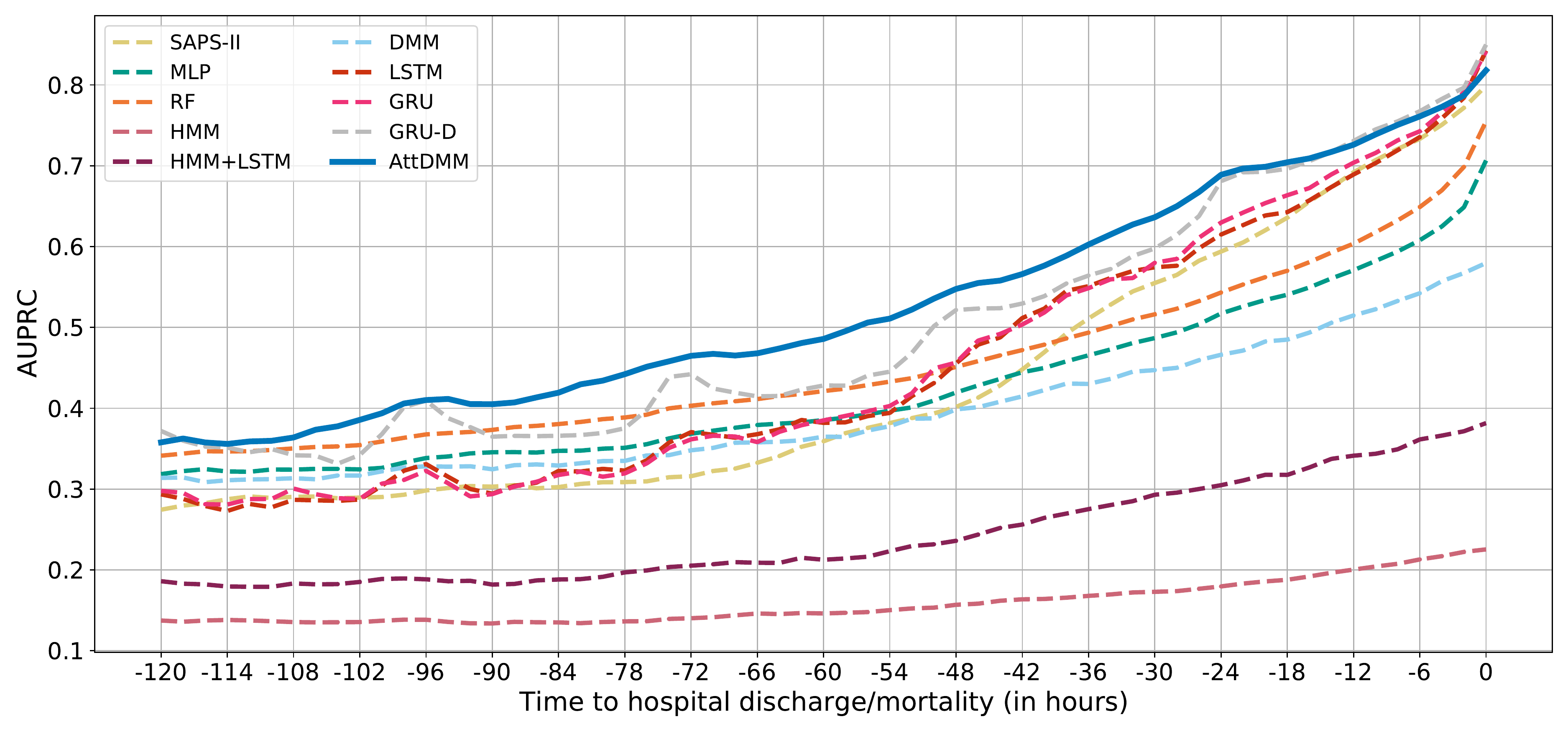}
        \label{fig:task2_auprc}
    \end{subfigure}
    
    \caption{Sensitivity of prediction performance across varying time windows, that is, when making predictions hours ahead of an ICU discharge/mortality. Shown are two performance metrics: AUROC (top) and AUPRC (bottom).}
    \label{fig:task2}
    
\end{figure}

Figure \ref{fig:task2} depicts the evolution of prediction performance across different time periods. For both the AUROC and AUPRC, \attdmm{} yields favorable results. In comparison, the baseline models vary in their performance at different steps. The random forest has an AUROC similar to \attdmm{} until 84 hours to discharge/death. After that, \attdmm{} outperforms the random forest. The ICU/GRU and ICU/GRU-D show increases in AUROC for the last 24 hours of hospital stays, but are ranked consistently below the \attdmm. For the AUPRC, \attdmm{} and the ICU/GRU-D show a similar performance for the last 24 hours of hospital stays. However, prior to that, the ICU/GRU-D and the other baselines are consistently outperformed by \attdmm. For both AUROC and AUPRC, the other sequential networks (\ie, ICU/LSTM, ICU/GRU, and ICU/GRU-D) show a lower performance level compared to \attdmm{} when focusing on the time window of more than 24 hours to discharge/death (\ie, from $-120$ hours to $-24$ hours). Overall, we see consistent gains from using \attdmm{} over existing baselines. 

\subsection{Example predictions} 
\label{sec:example_predictions}

The risk score is crucial for the decision-making of clinical practitioners in ICUs. It provides early warnings of the deterioration in a patient's health so that preventive measures can be taken. For this, our \attdmm{} outputs the probability of in-hospital mortality over time. This can inform physicians in ICU regarding the risk scores of each patient and, thus, influence the treatment. 

\begin{figure}[h]
    \begin{subfigure}{1.\columnwidth}
        \centering
        \includegraphics[width=1.\columnwidth]{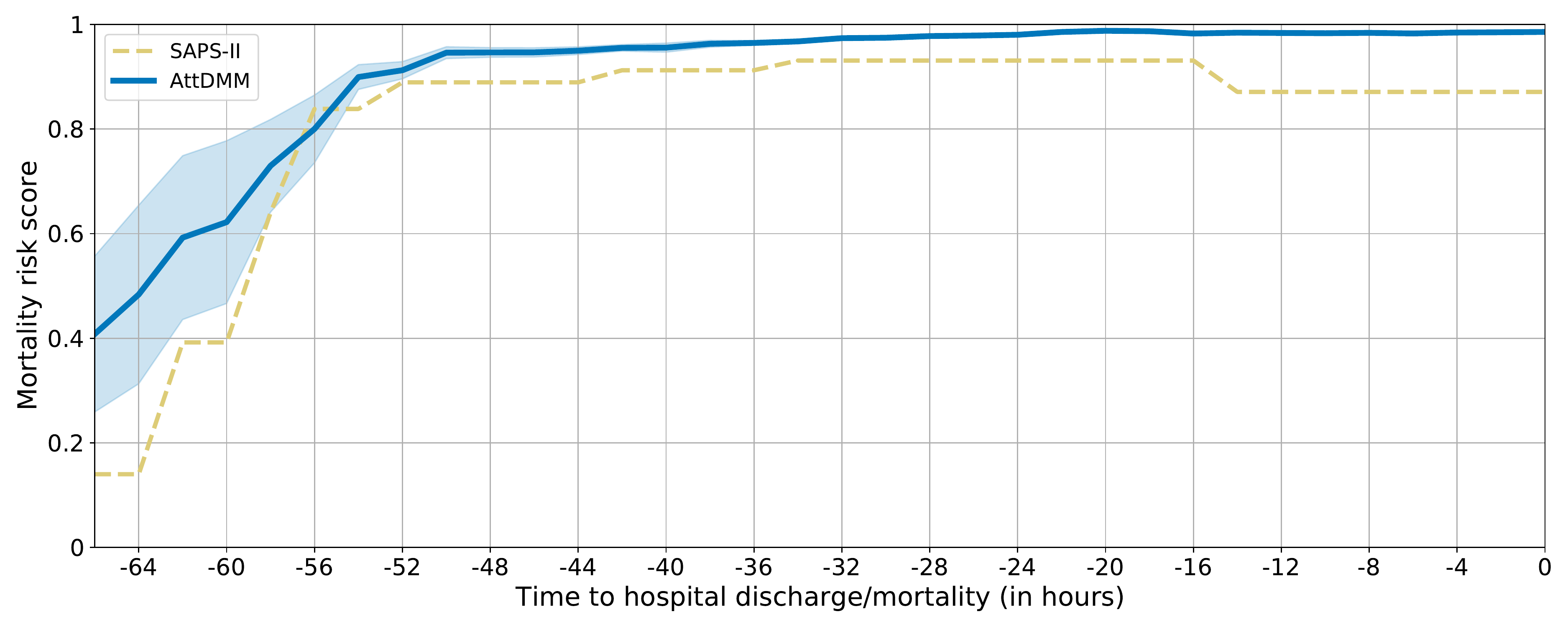}
        \label{fig:risk_death}
    \end{subfigure}
    
    \begin{subfigure}{1.\columnwidth}
        \centering
        \includegraphics[width=1.\columnwidth]{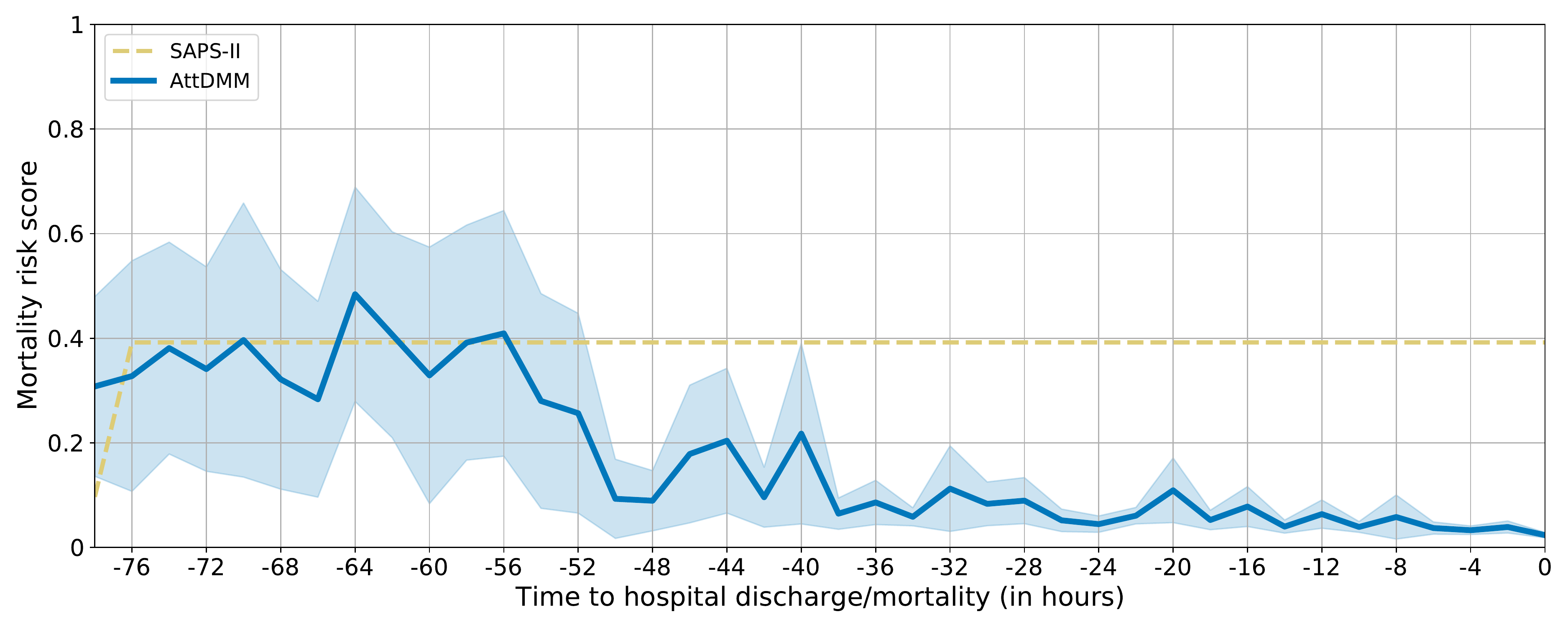}
        \label{fig:risk_discharge}
    \end{subfigure}
    
    \caption{Evolution of risk scores over time for two example patients, namely a patient with in-hospital mortality (top) and a patient with hospital discharge (bottom).}
    \label{fig:interpretability_risk_score}
    
\end{figure}

Figure \ref{fig:interpretability_risk_score} presents risk scores for two example patients, namely one example patient with in-hospital mortality (top) and one example patient with a hospital discharge (bottom). As shown by both patients, \attdmm produces proper risk scores several hours before death or discharge. For the patient who died in the hospital, \attdmm{} outputs a risk score that peaks quickly, thereby informing that the patient is at high risk. In this case, the risk score from \attdmm{} exceeds a mortality probability of 90\,\% and this risk score is retrieved even 52 hours before death. For the patient who has been discharged from the hospital, the risk scores produced by \attdmm{} suggests that the patient had a moderate risk of mortality during the first 30 hours of the ICU stay. After that, the health condition stabilized further, as is shown by the fact that the associated risk score approaches zero. In contrast, for the second patient, the clinical practice (\sapsii) estimates a fairly constant risk throughout the entire hospital stay. As \sapsii is computed based on the most critical measurements, it cannot properly model the temporal changes when the measurements are indicating a better health condition. Therefore, \sapsii does not capture such recovery in the health condition of the patient. 

\section{Discussion}

\textbf{Theory-informed model:} We present a novel generative deep probabilistic model for predicting mortality risk in ICUs. Our \attdmm outperformed existing baselines due to the way the model is formalized. Specifically, our \attdmm is modeled so that it jointly captures (1)~long-term disease dynamics (via an attention network) and (2)~different latent disease states in patient trajectories (via a latent variable model). By combining these two characteristics, \attdmm presents a powerful tool for modeling health progression in ICUs, as was confirmed in a superior prediction performance. 

\textbf{Clinical relevance:} \attdmm outperformed both baselines from clinical practice (\eg, \sapsii{} \citep{le1993new}) and state-of-the-art machine learning for ICU mortality prediction \citep{de2018deep, che2018recurrent, ge2018interpretable}. In practice, such an improvement in performance allows the triggering of warnings when health conditions deteriorate several hours earlier. Thereby, our model provides ICU practitioners with more time for early intervention in order to save patient lives.

\textbf{Precision of real-time risk score:} The real-time risk score provided by \attdmm{} provides the health trajectory of an ICU patient over time. Thereby, ICU practitioners can track the course of the disease, and, by identifying patients at risk, adapt their decision-making concerning treatment planning. Owing to our probabilistic setting, \attdmm{} additionally outputs the confidence interval of the predicted risk score. Thus, ICU practitioners are informed about the precision of the prediction and they might  postpone a discharge until more measurements are available. This is a particular benefit over risk scoring (\eg, SAPS, APACHE) in clinical practice, in which similar confidence intervals are lacking. 

\textbf{Generalizability:} We developed \attdmm{} as a generative deep probabilistic model tailored for ICU mortality. Needless to say, \attdmm{} is not limited to an ICU setting and might facilitate other use cases, in which inferences from time-series have to be made that have long-term dependencies and are driven by latent dynamics, such as clickstream analytics or churn prediction.

\section{Conclusion}

In intensive care units (ICUs), patients are subject to constant monitoring in order to guide clinical decision-making. While state-of-the-art models can capture long-term dependencies, these cannot formally account for latent disease states in health trajectories. To fill this gap, we developed a novel probabilistic generative model, specifically, an attentive deep Markov model (\attdmm). To the best of our knowledge, this is the first attentive deep Markov model. \attdmm was co-developed with clinical researchers who emphasized the benefit of real-time risk scoring. In our numerical experiments, \attdmm yielded performance improvements in both AUROC and AUPRC over the state of the art by more than 2\,\%. Altogether, this enables more accurate risk scoring so that lives of patients at risk of mortality can be saved.

\bibliographystyle{ACM-Reference-Format-no-doi}
\balance 
\bibliography{references.bib}

\clearpage
\appendix

\section{Estimation procedure} \label{sec:estimation_procedure}

The true posterior of the latent variables $p(z_{1:T} \,\mid\, x_{1:T},\, m_{1:T},\, s)$ is computationally intractable. Because of this, we adopted stochastic variational inference and leverage the approximated posterior $q(z_{1:T} \,\mid\, x_{1:T},\, m_{1:T},\, s)$ as the proxy of the true posterior. Throughout the iterations, the approximated posterior is getting closer, in terms of Kullback-Leibler (KL) divergence, to the true posterior. This is achieved by maximizing the ELBO. For this, we use stochastic optimization via unbiased Monte Carlo estimates of the gradient, details are found in \citep{ranganath2014black}. For the ICU mortality prediction task, we incorporate the maximization of the ELBO into the loss function as a regularization term.  Overall, in the following, we show how to estimate \attdmm{} via an end-to-end training task by a tailored ELBO.

\textbf{Loss function:} The loss function of \attdmm{} is given by
\begin{equation} \label{eq:loss}
   \mathcal{L}(y, \hat{y}, x) = \ell(y, \hat{y}) - \alpha\,  \ELBO(x) ,
\end{equation}
where the two terms $\ell(y, \hat{y})$ and the ELBO are described below. 

\textbf{Cross-entropy loss:} The term $\ell(y, \hat{y})$ denotes weighted cross-entropy loss between the observed label $y$ and the corresponding mortality prediction $\hat{y}$. It is given by
\begin{equation}
    \ell(y, \hat{y}) = - \rho\, y\, \log{(\hat{y})} - (1-y)\, \log{(1 - \hat{y})}
\end{equation}
with a weight $\rho = \frac{|\{ y \in Y : y = \text{"discharge"} \}|}{|\{ y \in Y : y = \text{"death"} \}|} $ denoting the discharge-to-death ratio. Introducing such weight further helps in discriminating the minority class (\ie, in-hospital mortality) in the imbalanced dataset. 

\textbf{ELBO:} We denote the evidence lower bound via ELBO. It serves as the regularization term of the loss function. The strength of the regularization is parametrized by $\alpha$. The ELBO formulation is given by
\begin{equation}
    \begin{aligned}[b]
    & \ELBO(x) = \mathbb{E}_{q(z_{1:T} \,\mid\, x_{1:T},\, m_{1:T},\, s)}[\log{p(x_{1:T} \,\mid\, z_{1:T})}] \\ 
    & \ -  \KL(q(z_{1:T} \,\mid\, x_{1:T},\, m_{1:T},\, s) \,\mid\mid\, p(z_{1:T})) .
    \end{aligned}
\end{equation}
The expectation term denotes the expected log-likelihood of $x_{1:T}$ given the latent variables $z_{1:T}$. The KL-divergence measures the similarity between the posterior approximation and the prior formulation of the latent variables $z_{1:T}$.

In this ELBO formulation, $q(z_{1:T} \,\mid\, x_{1:T},\, m_{1:T},\, s)$ is decomposed into further terms, which are produced by the posterior approximation, given by
\begin{equation}
    q(z_{1:T} \,\mid\, x_{1:T},\, m_{1:T},\, s) = \prod_{t=1}^T  q(z_t \,\mid\, z_{t-1},\, s,\, x_{t:T},\, m_{t:T}) .
\end{equation}

Similarly, $p(x_{1:T} \,\mid\, z_{1:T})$ is decomposed into individual terms of $x_t$, which are produced by the emission network. Specifically, we can rewrite
\begin{equation}
    p(x_{1:T} \,\mid\, z_{1:T}) = \prod_{t=1}^T p(x_t \,\mid\, z_t) .
\end{equation}

We make further adjustments to handle missing measurements inside the vector $x_t$. For this, \attdmm is specified to train on actual measurements (\ie, ignoring imputed values), which is achieved by 
\begin{equation}
    p(x_t \,\mid\, z_t) = \prod_{i : m_{ti}=1} p(x_{ti} \,\mid\, z_t) .
\end{equation}

\textbf{Sampling:} Computation of the above ELBO is analytically intractable due to latent variables being modeled by a continuous distribution and the non-linearity encoded in \attdmm. Therefore, we leverage Monte Carlo sampling and, thereby, we estimate ELBO via stochastic variational inference based on sampled latent variables.

\section{Descriptive Statistics of \mimiciii{} } \label{sec:app_mimic}

\input{tables/appendix/features_table}

\begin{figure}[h]
    \centering
    \includegraphics[width=1.0\columnwidth]{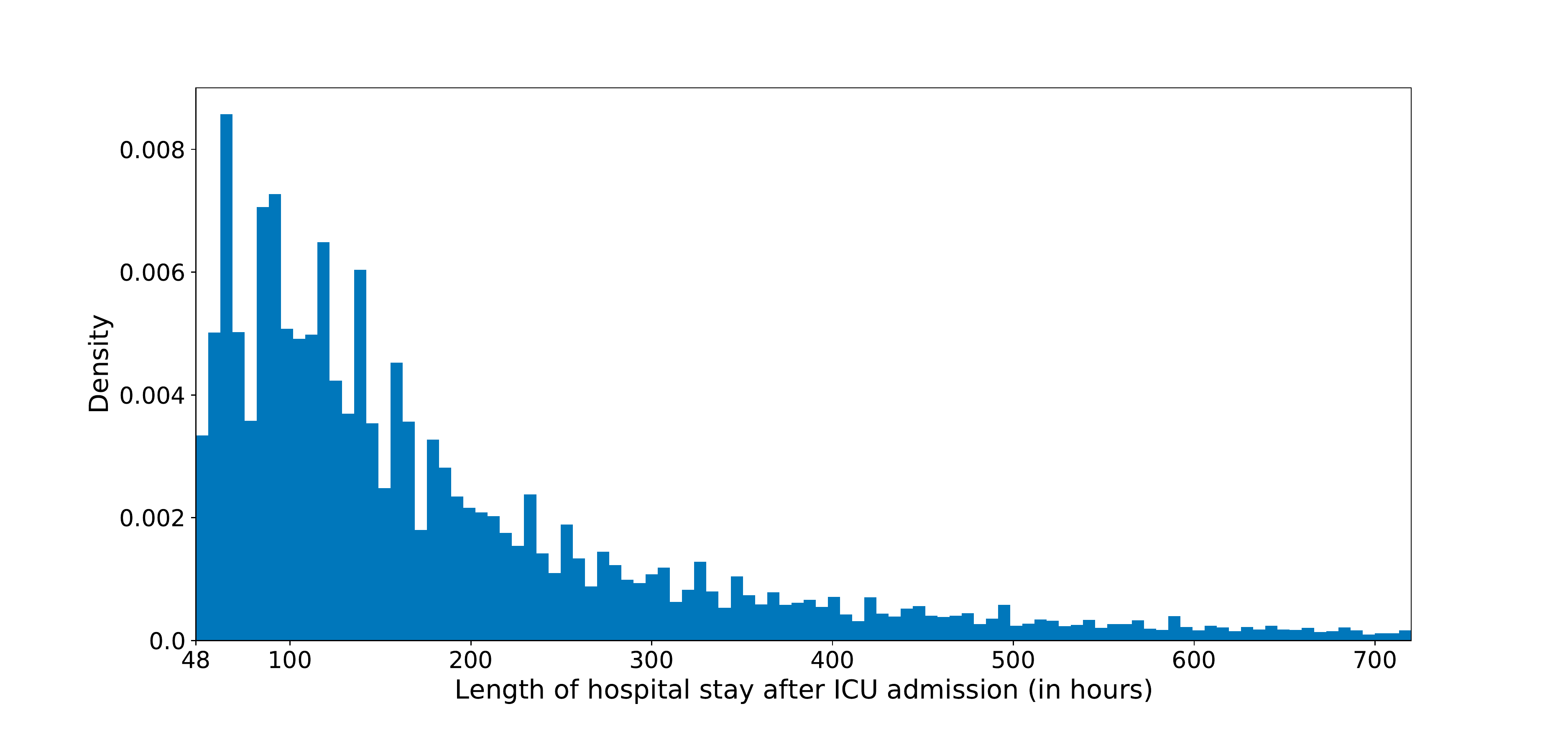}
    \caption{Histogram of length of hospital stay after ICU admission. Recall that hospital stays with less than 48 hours were removed during preprocessing as in \citep{purushotham2017benchmark, che2018recurrent}}.
    \label{fig:los_icu}
\end{figure}

\section{Implementation details} \label{sec:implementation_details}

The dataset is split randomly into 5 separate folds. Each fold has the same ratio of in-hospital mortality and hospital discharge. To quantify the uncertainty regions for the predictions, we used the following approach: each model was trained 5 times on each fold, in which one fold represents the test set and the remaining 4 folds were randomly split into a training set (3 folds) and a validation set (1 fold; for early stopping). Afterwards, the performance was aggregated over the different folds via macro averaging. Later, we report the standard deviation across folds.

The hyperparameters of the models are tuned according to performance on validation set. We use the hyperparameter set that provides the best overall validation score. Details on hyperparameter tuning are presented in Appendix~\ref{sec:model_parameters}.

All the models were implemented in Python. We  employed the following libraries: Pyro for \attdmm{}; PyTorch for the LSTM, GRU and MLP; Tensorflow for the GRU-D; and Scikit-learn for the random forest. Training and evaluations were performed on TITAN \rom{5} GPU from NVIDIA with 12 GB of memory. 

\section{Training Details} \label{sec:model_parameters}

Due to the large hyperparameter space across all the models, it was computationally expensive to perform a grid search. Therefore, we adopted a \textit{ceteris paribus} strategy which formally means that we tuned each parameter individually while keeping the other parameters fixed. We ran this procedure for a few loops, until we had observed the convergence of the score. The tuning parameters with the corresponding tuning ranges are listed for the \attdmm in Table~\ref{tab:attdmm_params} and for the baselines in Table~\ref{tab:baseline_params}. 

For both \attdmm and the baselines of the sequential neural network (ICU/LSTM, ICU/GRU, and ICU/GRU-D) we used the Adam optimizer. These models were trained using early stopping with patience set to 20 epochs.

\input{tables/appendix/attdmm_params}

\input{tables/appendix/baseline_params}

\end{document}

%% file: tables/experiments/experiments_48hours.tex
\begin{table}[h!]
\caption{Performance in prediction task 1 where in-hospital mortality risk is estimated at 48 hours after ICU admission.}
\label{tab:experiments_48hours}
\begin{center}
{\addtolength{\tabcolsep}{4pt} 
\begin{tabular}{p{0.35\columnwidth}p{0.2\columnwidth}p{0.2\columnwidth}}
\hline
\textbf{Model} & \multicolumn{1}{c}{\textbf{AUROC}} & \multicolumn{1}{c}{\textbf{AUPRC}} \\
\hline
\sapsii{} \citep{le1993new} & 0.747\,$\pm$\,0.007 & 0.255\,$\pm$\,0.014 \\
MLP & 0.832\,$\pm$\,0.014 & 0.395\,$\pm$\,0.027 \\
Random forest & 0.826\,$\pm$\,0.010 & 0.416\,$\pm$\,0.020 \\
HMM  & 0.711\,$\pm$\,0.007 & 0.173\,$\pm$\,0.003 \\
HMM+LSTM & 0.774\,$\pm$\,0.022 & 0.289\,$\pm$\,0.029 \\
DMM & 0.839\,$\pm$\,0.013 & 0.387\,$\pm$\,0.036 \\
ICU/LSTM \citep{ge2018interpretable} & 0.831\,$\pm$\,0.017 & 0.403\,$\pm$\,0.028 \\
ICU/GRU \citep{de2018deep} & 0.838\,$\pm$\,0.019 & 0.412\,$\pm$\,0.029 \\
ICU/GRU-D \citep{che2018recurrent} & 0.857\,$\pm$\,0.012 & 0.454\,$\pm$\,0.014 \\
\hline
\textbf{Proposed \emph{Att}DMM} & \textbf{0.876\,$\pm$\,0.004} & \textbf{0.465\,$\pm$\,0.010} \\
\hline
\multicolumn{3}{l}{\footnotesize Higher is better. Best value in bold.}
\end{tabular}
}
\end{center}
\end{table}

%% file: tables/experiments/experiments_cropped_t.tex
\begin{table}[h!]
\caption{Performance in prediction task 2 in which in-hospital mortality risk is estimated throughout the complete hospital stay.}
\label{tab:experiments_cropped_t}
\begin{center}
{\addtolength{\tabcolsep}{4pt} 
\begin{tabular}{p{0.35\columnwidth}p{0.2\columnwidth}p{0.2\columnwidth}}
\hline
\textbf{Model} & \multicolumn{1}{c}{\textbf{AUROC}} & \multicolumn{1}{c}{\textbf{AUPRC}} \\
\hline
\sapsii{} \citep{le1993new} & 0.829\,$\pm$\,0.008 & 0.449\,$\pm$\,0.014 \\
MLP & 0.821\,$\pm$\,0.013 & 0.439\,$\pm$\,0.022 \\
Random forest & 0.846\,$\pm$\,0.010 & 0.467\,$\pm$\,0.023\\
HMM & 0.672\,$\pm$\,0.032 & 0.161\,$\pm$\,0.012 \\
HMM+LSTM & 0.759\,$\pm$\,0.022 & 0.249\,$\pm$\,0.034 \\
DMM & 0.815\,$\pm$\,0.010 & 0.403 \,$\pm$\,0.026 \\
ICU/LSTM \citep{ge2018interpretable} & 0.829\,$\pm$\,0.012 & 0.470\,$\pm$\,0.026 \\
ICU/GRU \citep{de2018deep} & 0.842\,$\pm$\,0.016 & 0.473\,$\pm$\,0.034 \\
ICU/GRU-D \citep{che2018recurrent} & 0.834\,$\pm$\,0.016 & 0.517\,$\pm$\,0.025 \\
\hline
\textbf{Proposed \emph{Att}DMM} & \textbf{0.865\,$\pm$\,0.010} & \textbf{0.545\,$\pm$\,0.029} \\
\hline
\multicolumn{3}{l}{\footnotesize Higher is better. Best value in bold.}
\end{tabular}
}
\end{center}
\end{table}

%% file: tables/appendix/features_table.tex
\begin{table}[h!]
\caption{Features from \mimiciii{} used in predictions}
\label{tab:table_features}
\begin{center}
\begin{tabular}{p{0.4\columnwidth}p{0.25\columnwidth}S[table-format=2.2]}
\hline
\textbf{Feature}	& \textbf{Type} & \textbf{Missingness (\%)}\\
\hline
Glasgow coma scale & time-series & 70.04  \\
Systolic blood pressure & time-series & 49.36 \\
Heart rate & time-series & 48.69 \\
Body temperature & time-series & 70.46 \\
PaO$_2$/FiO$_2$ & time-series & 93.88 \\
Urinary output & time-series & 57.16 \\
Serum urea nitrogen level & time-series & 88.88 \\
White blood cells count & time-series & 89.84 \\
Serum bicarbonate level & time-series & 89.11 \\
Sodium level & time-series & 87.63 \\
Potassium level & time-series & 85.29 \\
Bilirubin level & time-series & 97.47 \\
Age & static & 0.00 \\
Acquired immunodeficiency syndrome & static & 0.00 \\
Hematologic malignancy & static & 0.00 \\
Metastatic cancer & static & 0.00 \\
Admission type & static & 0.00 \\
\hline
\end{tabular}
\end{center}
\end{table}

%% file: tables/appendix/attdmm_params.tex
\begin{table}[h!]
\caption{Hyperparameters used for tuning \attdmm.}
\label{tab:attdmm_params}
\begin{center}
\begin{tabular}{p{0.6\columnwidth}p{0.4\columnwidth}}
\hline
\textbf{Tuning parameters} & \textbf{Tuning range} \\
\hline
Dimension of latent variable & 8, 16, 24, 48 \\
Dimension of emission hidden layer & 16, 32, 48, 96 \\
Dimension of transition hidden layer & 32, 64, 96, 192 \\
Dimension of RNN cell & 8, 16, 24, 48 \\
Dimension of attention mechanism & 16, 24, 36, 48 \\
Dimension of MLP hidden layer & 2, 4, 8, 12 \\
Regularization strength of ELBO ($\alpha$) & 0.001, 0.01, 0.1, 1 \\
Number of Monte Carlo samples ($N$) & 1, 2, 5, 10, 20 \\
Learning rate & 0.00002, 0.0002, 0.001, 0.01 \\
Batch size & 64, 128, 256 \\
\hline
\end{tabular}
\end{center}
\end{table}

%% file: tables/appendix/baseline_params.tex
\begin{table}[h!]
\caption{Hyperparameters used for tuning the baseline models.}
\label{tab:baseline_params}
\begin{center}
\begin{tabular}{p{0.2\columnwidth}p{0.4\columnwidth}p{0.25\columnwidth}}
\hline
\textbf{Model} & \textbf{Tuning parameters} & \textbf{Tuning range} \\
\hline
MLP & Dimension of hidden layer & 2, 4, 8, 12 \\
 & Number of hidden layers & 1, 2, 3 \\
 & Learning rate & 0.00002, 0.0002, 0.001, 0.01 \\
 & Batch size & 64, 128, 256 \\
\hline
RF & Number of trees & 50, 100, 500, 1000 \\
 & Maximum depth & 3, 5, 7, 9, 11 \\
 & Class weight & none, balanced \\
\hline
HMM+LSTM* & Number of states & 2, 3, ..., 10 \\
\hline
ICU/LSTM, & Dimension of cell state & 8, 16, 24, 48 \\
ICU/GRU, & Dropout rate & 0, 0.1, 0.2, 0.5 \\
ICU/GRU-D& Learning rate & 0.00002, 0.0002, 0.001, 0.01 \\
 & Batch size & 64, 128, 256 \\
\hline
\multicolumn{3}{l}{\footnotesize * LSTM of HMM+LSTM is tuned based on the same parameter space of ICU/LSTM.}
\end{tabular}
\end{center}
\end{table}